# Implementing a new fully stepwise decomposition-based sampling technique for the hybrid water level forecasting model in real-world application


Ziqian Zhang[a], Nana Bao[b,*], Xingting Yan[c,*], Aokai Zhu[d], Chenyang Li[b], Mingyu Liu[b]

[a] School of Cyber Science and Technology, University of Science and Technology of China, Hefei 230026, PR China

[b] School of Internet, Anhui University, Hefei 230039, PR China

[c] Chinese Academy of Sciences, Institute of Plasma Physics, Hefei 230031, PR China

[d] School of Software Technology, Zhejiang University, Ningbo 315048, PR China

* Corresponding authors.

E-mail address: nnbao@ahu.edu.cn (N. Bao), xingting.yan@ipp.ac.cn (X. Yan).


## Abstract:


Various time variant non-stationary signals need to be pre-processed properly in hydrological time series forecasting in real world, for example, predictions of water level. Decomposition method is a good candidate and widely used in such a pre-processing problem. However, decomposition methods with an inappropriate sampling technique may introduce future data which is not available in practical applications, and result in incorrect decomposition-based forecasting models. In this work, a novel Fully Stepwise Decomposition-Based (FSDB) sampling technique is well designed for the decomposition-based forecasting model, strictly avoiding introducing future information. This sampling technique with decomposition methods, such as Variational Mode Decomposition (VMD) and Singular spectrum analysis (SSA), is applied to predict water level time series in three different stations of Guoyang and Chaohu basins in China. Results of VMD-based hybrid model using FSDB sampling




technique show that Nash-Sutcliffe Efficiency (NSE) coefficient is increased by 6.4%, 28.8% and 7.0% in three stations respectively, compared with those obtained from the currently most advanced sampling technique. In the meantime, for series of SSA-based experiments, NSE is increased by 3.2%, 3.1% and 1.1% respectively. We conclude that the newly developed FSDB sampling technique can be used to enhance the performance of decomposition-based hybrid model in water level time series forecasting in real world.



# 1. Introduction

Reliable and accurate water level forecasting is critical for various activities in management and planning of water resources, such as hydropower generation, water supplies and flood mitigation. However, due to climate change and human activities, water level time series, as one representative kind of hydrological time series, have become increasingly nonlinear, stochastic and complex, which makes precise water level forecasting more difficult (Dodov and Foufoula-Georgiou, 2005; Wang et al., 2006; Zhu et al., 2019) (Dodov and Foufoula-Georgiou, 2005; Wang et al., 2006; Zhu et al., 2019). It is necessary to develop an effective and accurate water level forecasting model to avoid above nonlinear and non-stationary characteristics of signals.



Over the past decades, various water level forecasting models have been proposed and improved. These models can be divided into physical models (Cao et al., 2007; Clark and Hay, 2004; Singh and Sankarasubramanian, 2014; Yuan and Wood, 2012) and data-driven models (Chitsaz et al., 2016; Ghaith et al., 2020; Osman et al., 2020; Quilty et al., 2019). Physical models have outstanding interpretability, which help to clarify the fundamental physical process of hydrological forecasting (Cui et al., 2021). However, these physical models need a lot of complex parameter calibration work, which relies heavily on the expert's experience knowledge, difficult to be used in practice. In contrast, with large historical hydrological records accumulated, data-driven model has becoming simple and easy to apply. Besides, these data-driven models have the advantages of capturing empirical relationship between explanatory variables and response variables. The flexibility and practicability of this kind of model make it widely used in hydrological forecasting work. (Bruen and Yang, 2006; Ni et al., 2020; Xu et al., 2022). Statistical models and machine learning (ML) models are two kinds of data-driven models in time series forecasting work. Statistical models, such as autoregression (AR), moving average (MA), autoregressive moving average (ARMA), autoregressive integrated moving average (ARIMA), can model water level series easily only with historical water level data. (Chua and Wong, 2011; Jabbari and Bae, 2020; Jiang et al., 2018; Wang et al., 2018). The climate change and the intensifying human activity cause the observable hydrological data becoming more nonlinear and non-stationary. These statistical models cannot fit the increasing nonlinear and non-stationary relationships involved in hydrological processes well, now that they are essentially linear models for stationary data (Zhang et al., 2015). To solve this, various ML models have been introduced and widely used for hydrological time series forecasting, such as artificial neural networks (ANNs), support vector regression (SVR), recurrent



neural networks (RNNs), (Adhikary et al., 2018; Ahmed et al., 2021; Erdal and Karakurt, 2013).

Among these models, RNN is one of the most commonly used models because they are extremely

adept at handling time series data. But the forecasting accuracy of RNNs is still far from satisfactory

due to the gradient disappearance and gradient explosion during network training. To overcome

these problems of RNNs, Long short-term memory (LSTM) (Hochreiter and Schmidhuber, 1997)

is proposed to analyze long-term time series by using memory unit and three different gating units,

which has been widely used in hydrological forecasting (Liang et al., 2020; Liu et al., 2021; Zhang

et al., 2022). But, it is difficult to apply LSTM model in real time because the model needs huge

matrix calculations and sufficient iterations to ensure the minimum training error (Cui et al., 2021).

Ensemble learning models such as Extreme Gradient Boosting (XGBoost) (Chen and Guestrin, 2016)

can also be a good choice for hydrological forecasting problem. Compared with LSTM, XGBoost

is a fast, efficient and scalable model. It supports efficient parallel training, which dramatically

reduces the computational time of model training and validating while ensuring satisfactory

performance (Cui et al., 2021).

Even so, ML models may ignore the inherently nonlinear, complex and non-stationary

characteristics of hydrological signals, unable to adequately extract data features if pre/post-

processing of the input/output is not performed (Cannas et al., 2006; Zhang et al., 2015). Series

decomposition methods is an accessible pre-process technique to improve ML model performance,

by resolving a non-stationary time series into a set of components. So that, ML models can extract

simpler patterns from important components and aggregate them to train the final forecasting model.

This kind of models are called decomposition-based hybrid models (Shi et al., 2021). These hybrid

models have increasingly studied in the literature for various time series forecasting problems,



including water level forecasting (Hu et al., 2021; Sibtain et al., 2020; Tayyab et al., 2018; Yu et al., 2018). Commonly used decomposition methods include Discrete Wavelet Transform (DWT), Empirical Mode Decomposition (EMD), Ensemble Empirical Mode Decomposition (EEMD), Variational Mode Decomposition (VMD) and Singular Spectrum Analysis (SSA). Hu et al. (2021) propose a VMD-BPNN decomposition-based model to extract multiscale flow features combined VMD and Back-Propagation Neural Networks (BPNN). The results confirm that VMD-BPNN outperforms a single BPNN model without any decomposition. Tayyab et al., (2018) propose hybrid NN models combined DWT and EEMD, respectively. Result shows that forecasting accuracies of two decomposition-based models are better than those using ML model only. Wang et al. (2019) propose a data-driven model based on a double-processing strategy, combing with three decomposition methods, i.e., SSA, Improved Complete Ensemble Empirical Mode Decomposition with Adaptive Noise (ICEEMDAN) and Extreme Learning Machine (ELM). Good results also obtained.

Recently, the sampling technique in decomposition method has attracted the attention of researchers (Du et al., 2017; Fang et al., 2019; Karthikeyan and Nagesh Kumar, 2013; Xu et al., 2022). Xu et al. (2022) mentions that decomposition methods with an inappropriate sampling technique may introduce future data which is not available in practical applications, and result in information leakage and spuriously high forecasting accuracy. The widely used Overall Decomposition-Based (ODB) sampling technique is one of them. ODB sampling technique first decomposes the entire water level series and then divide the decomposition components into calibration and validation sets to establish models. However, the explanatory variables of all samples are actually computed using information on future data that is unavailable in the current



time step (Napolitano et al., 2011). Therefore, the models developed using ODB sampling technique typically lead to misleading and overestimated performance that cannot be practically applied in the real-world water level forecasting. Few studies have focused on such a problem, but it is delighted that there are still some studies having noted the incorrect or impractical usage of sampling techniques and proposed possible solutions. Karthikeyan and Nagesh Kumar (2013) propose an improved sampling technique named KN (K for Karthikeyan and N for Nagesh Kumar). The KN sampling technique divides original series into calibration period and validation period. Data during the former one will be decomposed like ODB sampling technique, and the others are sequentially appended to the calibration period for stepwise decomposition like actual forecasting practice. However, studies have confirmed that this sampling technique outperforms a single model in the calibration period, but does not achieve satisfactory forecasting performance in the validation period (Fang et al., 2019; He et al., 2022). To maintain the consistent model performance, Fang et al. (2019) propose a Stepwise Decomposition-Based (SDB) sampling technique, strictly following stepwise decomposition strategy to extract explanatory variables of calibration and validation periods. First, an initial series is selected for decomposition to extract the first explanatory variables; next, the newly observed data are appended to the initial series in turn for stepwise decomposition until all observed data is decomposed and explanatory variables are extracted. SDB sampling technique excludes the influence of future information on explanatory variables. However, as with ODB and KN, SDB decomposes the entire series to obtain the response variables of all samples. This leads to biased forecasting targets which makes it difficult to achieve satisfactory results. Thus, using appropriate sampling techniques to obtain response variables is particularly important to improve the forecasting performance of decomposition-based hybrid models. This paper proposes an



improved Fully Stepwise Decomposition-Based (FSDB) sampling technique considering response variables correction.

The main contributions of this paper can be summarized as follows: (1) proposing an improved Fully Stepwise Decomposition-Based (FSDB) sampling technique considering response variables correction; (2) taking the daily water level forecasting of the Guoyang and Chaohu basins in Anhui province, China as an example, and verify the rationality and superiority of the proposed method by comparing with SDB sampling technique;

The rest of this paper is organized as follows. In Section 2, the develop process for decomposition-based hybrid model and feasible sampling techniques are introduced. In Section 3, the case application and evaluation metrics are presented. In Section 4, various experiments are carried out to test the FSDB sampling technique using two decomposition methods, where results are analyzed and summarized. Section 5 summarizes this work, and suggests an improvement for future research.

**Table 1**

List of symbols

| | |
|---|---|
| AR | Autoregression |
| MA | Moving average |
| ARMA | Autoregressive moving average |
| ARIMA | Autoregressive integrated moving average |
| ML | Machine learning |
| ANNs | Artificial neural networks |
| SVR | Support vector regression |
| RNNs | Recurrent neural networks |
| LSTM | Long short-term memory |
| DWT | Discrete wavelet transform |
| EMD | Empirical mode decomposition |
| VMD | Variational mode decomposition |
| SSA | Singular spectrum analysis |
| BPNN | Back-propagation neural networks |
| EEMD | Ensemble empirical mode decomposition |



| ICEEMDAN | Improved complete ensemble empirical mode decomposition with adaptive noise |
| --- | --- |
| ELM | Extreme learning machine |
| ODB | Overall decomposition-based |
| KN | Karthikeyan and Nagesh Kumar sampling technique |
| SDB | Stepwise decomposition-based sampling technique |
| FSDB | Fully stepwise decomposition-based sampling technique |
| IMFs | Intrinsic mode functions |
| XGBoost | Extreme Gradient Boosting |
| NSE | Nash-Sutcliffe efficiency coefficient |
| RMSE | Root mean square error |
| MAE | Mean absolute error |
| n_estimators | Number of decision trees in XGBoost |
| learning_rate | Learning rate, controlling the step size at each iteration when updating the weights |
| max_depth | Depth of decision tree |
| min_child_weight | Minimum sum of instance weight needed in a leaf node |
| DL | Decomposition level |

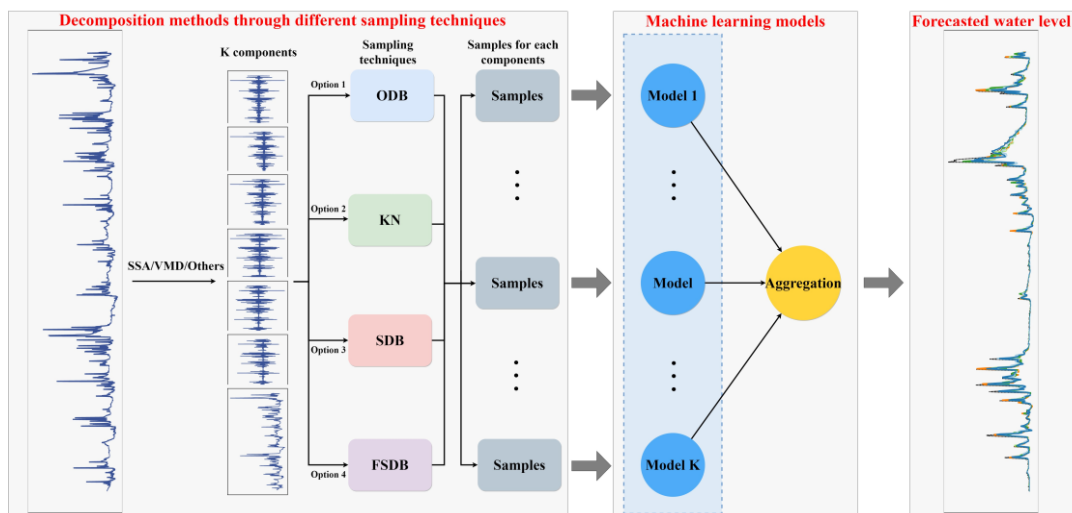

**Fig. 1.** Development process for decomposition-based hybrid models



# 2. Decomposition-based Model using FSDB sampling technique

This section introduces the newly developed FSDB sampling technique based on the structure of decomposition-based hybrid models. Section 2.1 describes detailed development process for decomposition-based hybrid models following decomposition-forecasting-aggregation principle (Fang et al., 2019). Section 2.2 presents the FSDB sampling technique after the introduction of three common sampling techniques.

## 2.1 Development process for decomposition-based hybrid models

The development process for decomposition-based hybrid models is shown in Fig. 1. First, the original non-stationary time series can be decomposed into a set of components (i.e., $K$ components in Fig.1.) by decomposition method such as SSA and VMD. Meanwhile, $K$ components need to be processed into some samples respectively, using sampling technique, such as ODB, KN, SDB. Next, samples of one component can be used as training data for developing a separate basic ML model. Finally, predicted value of these single ML models can be aggregated for the final water level forecasting.

### 2.1.1 Decomposition Methods through different sampling techniques

To determine whether the investigated sampling techniques are generally applicable to decomposition-based hybrid model, different types of decomposition methods are used in this paper. SSA and VMD, as the two widely used methods among these familiar decomposition methods (such as EMD, VMD, EEMD, DWT, SSA), are used as pre-processing methods for water level series.



(1) SSA is a relatively new method of for time series analysis and forecasting time series, which integrates time series analysis, multivariate statistics, multivariate geometry, dynamical systems and signal processing (Chen et al., 2022). The central idea of SSA is to decompose the original time series totally into the sum of its independent components, which can be identified as trends, periodic oscillation, and noise (Rocco, 2013). The original input series can be obtained by simply summing the components. The main steps of SSA include embedding, singular value decomposition, grouping and diagonalization averaging (Hassani, 2007). Compared with other decomposition methods, SSA has the advantages of simplicity and efficiency, which only needs to predetermine in which only one parameter--the length of embedding window (i.e., the number of components, namely), needs to be predetermined.

(2) VMD is a widely used and powerful decomposition method that can decompose data at multiple resolution levels combining with Wiener filtering, Hilbert transform and Alternating Direction Method of Multipliers (Dragomiretskiy and Zosso, 2014; Meng et al., 2021). The method combines Wiener filtering, Hilbert transform and Alternating Direction Method of Multipliers (Dragomiretskiy and Zosso, 2014). The VMD algorithm decomposes the input signal s(t) into $K$ intrinsic mode functions (IMFs). Two parameters of VMD, namely the decomposition level $K$ and the penalty factor used to balance the data-fidelity constrain $\alpha$, should be predetermined (Upadhyay and Pachori, 2015). $K$ determines the decomposition level of the data series, needing an appropriate parameter value. If the value is too small, poor extraction of IMFs from the input signal will be obtained, and vice versa. a too-small value of $K$ may lead to poor extraction of IMFs from the input signal, a too-large value of $K$ will cause the original series to be over-decomposed, where several IMFs share one center frequency, resulting in mode mixing



or frequency mixing. The value of $\alpha$ is inversely proportional to the bandwidths of the IMFs., also needing to be defined reasonably. If the value is too small, a larger bandwidth, information redundancy and additional noise for the IMFs will be generated, and vice versa. A too-small value of $\alpha$ may result in a larger bandwidth, information redundancy and additional noise for the IMFs. A too-large value of $\alpha$ may result in smaller bandwidths of the IMFs and thus loss of some information of the original series (Zuo et al., 2020). It is worth noting that in contrast to SSA, VMD decomposition method suffers from reconstruction error, which means that the obtained components cannot accurately reconstruct the original input series. The decomposition results of VMD are sensitive to the decomposition level $K$ and the penalty factor $\alpha$. In this study, various combinations of $K$ and $\alpha$ were tested to obtain the optimal one. More details about SSA and VMD can be found in Hassani (2007) and Dragomiretskiy and Zosso (2014).

## 2.1.2 Machine learning model for forecasting

XGBoost has been applied to a variety of forecasting tasks with superior performance (Chen et al., 2021; Nguyen et al., 2021; Singha et al., 2021; Wu et al., 2022). Hence, it is chosen as the baseline model to forecast water level data in this study.

XGBoost is an ensemble model of decision tree with forward stepwise algorithm to seek the optimal value, and can be used for classification and regression. It was proposed by Chen and Guestrin (2016). It is an additive model consisting of $t$ trees, which takes the residual between the model predicted (the $t-1$ tree) and actual values as the prediction target for the next tree (the $t$ tree), as shown in Eq. (1).

$$\hat{y}^{(t)} = \sum_{j=1}^{t} f_j(x_i) = \hat{y}_i^{(t-1)} + f_t(x_i) \tag{1}$$



Where $x_i$ and $\hat{y}_i^{(t-1)}$ denotes the $i$ -th sample and the predicted value of the $t-1$ tree, $f_t(x_i)$ represents the prediction of the residual between the predicted value of the $t-1$ tree and the actual value. Together, they form the final prediction of the $t$ decision tree for the $i$ -th sample, namely $\hat{y}^{(t)}$.

In order to reduce the risk of overfitting, XGBoost introduces a regular term, which acts as a penalty term to control the complexity of the model. The formula is as follows:

$$\Omega(f_t) = \gamma T + \frac{1}{2}\lambda \sum_{j=1}^{T} \omega_j^2 \tag{2}$$

Where $\gamma$ and $\lambda$ control the strength of the penalty, $T$ is the number of leaf nodes, and $\omega$ is the weight.

As a supervised ML algorithm, XGBoost has an objective function including loss function $L$, penalty term $\Omega$ and a *constant*, defined in Eq. (3):

$$obj^{(t)} = \sum_{i=1}^{n} L(y_i, \hat{y}_i^{(t-1)} + f_t(x_i)) + \Omega(f_t) + constant \tag{3}$$

XGBoost performs a second-order Taylor expansion of the objective function $obj^t$ at $\hat{y}_i^{t-1}$. The expanded equation is shown below:

$$obj^{(t)} = \sum_{i=1}^{n} \left( g_i f_t(x_i) + \frac{1}{2} h_i f_t^2(x_i) \right) + \Omega(f_t) + constant \tag{4}$$

$$g_i = \partial_{\hat{y}^{(t-1)}} L(y_i, \hat{y}_i^{(t-1)}), h_i = \partial_{\hat{y}_i^{(t-1)}}^2 L(y_i, \hat{y}_t^{(t-1)}) \tag{5}$$

By removing all the constant terms, the simplified form is shown in Eq. (6):

$$obj^{(t)} = \sum_{i=1}^{n} \left( g_i f_t(x_i) + \frac{1}{2} h_i f_t^2(x_i) \right) + \Omega(f_t) \tag{6}$$



## 2.2 Feasible sampling techniques

### 2.2.1 Three existing sampling techniques

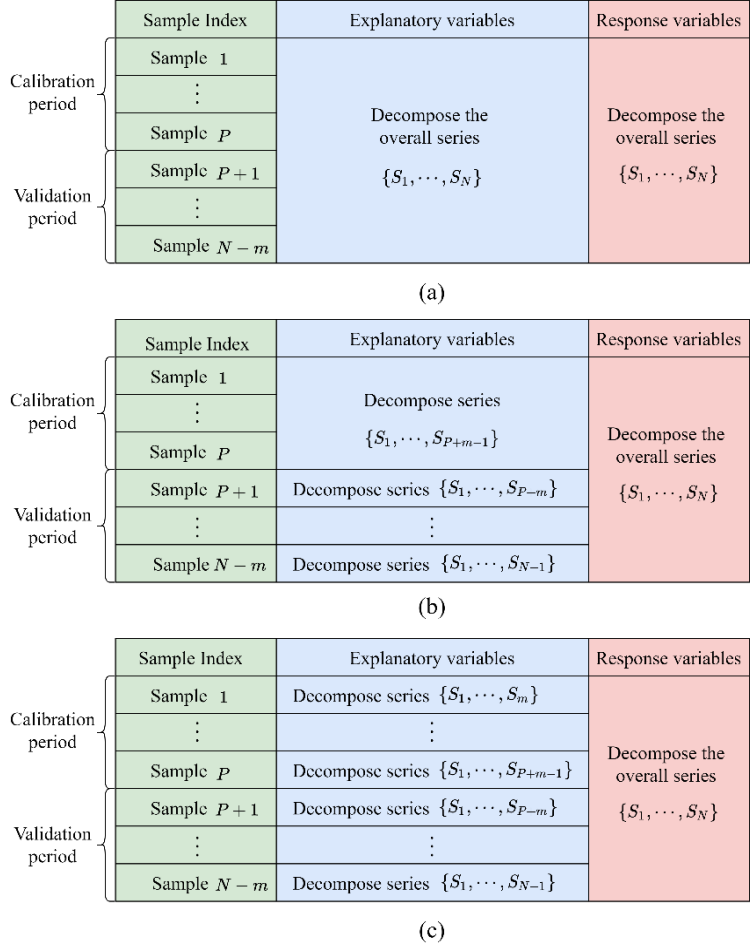

(a)

(b)

(c)

**Fig. 2.** Three existing sampling techniques used to extract explanatory variables and response variables of each sample. (a) ODB sampling technique; (b) KN sampling technique; (c) SDB sampling technique.

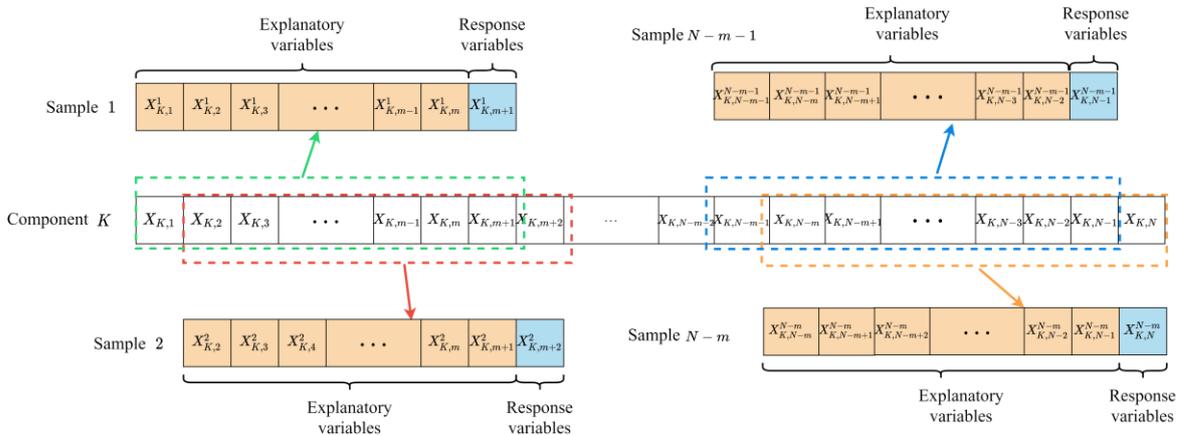

**Fig. 3.** Samples extraction from component $K$ using the sliding window method when the lag time is $m$.

Fig. 2 depicts the basic processes of three existing sampling techniques. The response variables in three sampling techniques are all extracted by decomposing the whole series in the



same way, but the explanatory variables are extracted differently. Assume that $S$ is a water level series of length $N$. $S_i$ is the $i$-th water level record in the series. The $m$ past observations before the water level to be forecasted are used as explanatory variables. In this paper, $m$ is chosen as 7.

As a mostly used sampling technique, the ODB method (shown in Fig. 2(a)) can be illustrated as follows:

(1) The overall original series $\{S_1, S_2, \ldots, S_N\}$ is decomposed into $K$ components using decomposition methods such as SSA or VMD.

(2) For each component, all of samples for the calibration and validation periods are extracted using the sliding window method. Fig 3 shows the sample extraction process when the lag time is $m$.

(3) Separate models are trained to fit samples in calibration period for each component, respectively.

(4) The explanatory variables of the validation period are sent to the forecasting model to produce forecasts of the response variables corresponding to each component. The final forecasted values are obtained by aggregating the forecasts of response variables for all components.

The ODB sampling technique extracts all samples by decomposing the whole original series, which is equivalent to assuming that the future data are known in each time step. But in practical applications, it is impossible to know the future data at the forecasting stage. Du et al. (2017) has shown that future data can have an impact on the decomposition results of previous data, which makes the explanatory variables of the samples obtained using the ODB sampling



technique contain additional information about the future data. Therefore, a stepwise decomposition strategy should be adopted to exclude the influence of future data on the forecasting performance and credibility of the model. Karthikeyan and Nagesh Kumar (2013) proposed a sampling technique that is more suitable for practical applications, namely the KN sampling technique.

The KN sampling technique shown in Fig. 2(b) can be explained as follows:

(1) The original series $\{S_1, S_2, \ldots, S_N\}$ is divided into calibration period and validation period first.

(2) The entire calibration period is decomposed to extract the explanatory variables of all of the samples in the calibration period. To acquire the explanatory variables of samples in the validation period, data from the validation period are sequentially appended to the calibration series. Once a new data is appended, the extended series will be decomposed to extract the explanatory variables for one sample, until all of the data in validation period have been appended to the calibration series and the explanatory variables are all extracted.

For calibration and validation periods, KN sampling technique follows the ODB sampling technique and the actual forecasting practice to extract samples separately. It has been shown that KN sampling technique has a high model performance in the calibration period, but does not achieve satisfactory forecasting performance in the validation period.

To maintain the consistent model performance throughout the calibration period, validation period and actual forecasting practice, Fang et al. (2019) proposes an SDB sampling technique that



strictly follows the stepwise decomposition strategy to extract all calibration and validation samples.

The SDB sampling technique shown in Fig. 2(c) is explained as follows:

(1) The series segment $\{S_1, S_2, \ldots, S_m\}$ is decomposed into $K$ components, and the last $m$ elements of each component are extracted as explanatory variables of the first sample, i.e. $\{X_{k,1}^1, X_{k,2}^1, \ldots, X_{k,m}^1\}, k \in \{1, 2, \ldots, K\}$ ,. To obtain the explanatory variables of the second sample, the observed data $S_{m+1}$ is appended to $\{S_1, S_2, \ldots, S_m\}$ and the extended series segment $\{S_1, S_2, \ldots, S_m, S_{m+1}\}$ is decomposed. The last $m$ elements of each component are extracted as explanatory variables of the second sample, i.e. $\{X_{k,2}^2, X_{k,3}^2, \ldots, X_{k,m+1}^2\}, k \in \{1, 2, \ldots, K\}$ . The series segment is gradually extended by appending the newly observed data one by one and then the explanatory variables of corresponding samples are extracted. This process continues until all the necessary explanatory variables are extracted.

(2) Finally, all samples are split into calibration and validation periods.

## 2.2.2 Fully Stepwise Decomposition-Based (FSDB) sampling technique

The ODB, KN and SDB sampling techniques decompose the overall series to extract response variables. This leads to biased forecasting targets which makes it difficult to achieve satisfactory model performance. To obtain response variables more reasonably, this study proposes a fully stepwise decomposition-based sampling technique considering response variables correction, which is shown in Fig. 4.



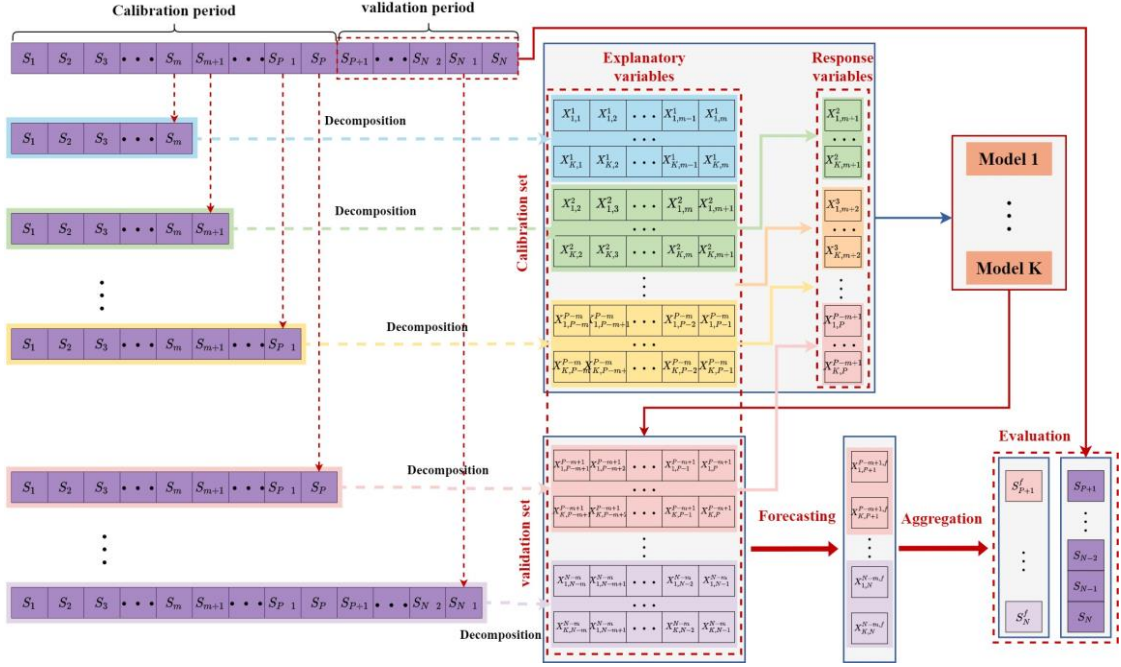

**Fig. 4.** The process of proposed FSDB sampling technique used to extract explanatory variables and response variables of each sample.

The explanatory variables in FSDB technique are extracted in the same way as SDB technique, except that the response variable of each sample is set to be the last explanatory variable of the next sample. For example, to obtain the explanatory variables of sample $M$, the series segment $\{S_1, S_2, \ldots, S_{M+m-1}\}$ is decomposed into $K$ components. The last $m$ elements of each component $\{X_{k,M}^M, X_{k,M+1}^M, \ldots, X_{k,M+m-1}^M\}, k \in \{1, 2, \ldots, K\}$ are extracted to constitute explanatory variables of sample $M$. Meanwhile, $\{X_{k,M+m-1}^M\}, k \in \{1, 2, \ldots, K\}$ are used as the response variables of the previous sample, i.e. sample $M-1$. Once all the explanatory and response variables are obtained, all the samples are split into calibration and validation periods.

The FSDB sampling technique strictly follows the stepwise decomposition strategy to generate explanatory variables for all samples, and the information of future data can be strictly excluded, which is the same as SDB technique. In addition, FSDB sampling technique does not simply decompose the overall time series to obtain the response variables of calibration period and verification period like ODB, KN and SDB sampling techniques, but obtains the response variables of the previous sample by extracting the explanatory variables of the current sample. This allows the produced response variables to be unaffected by future data and avoid biased forecasting targets.



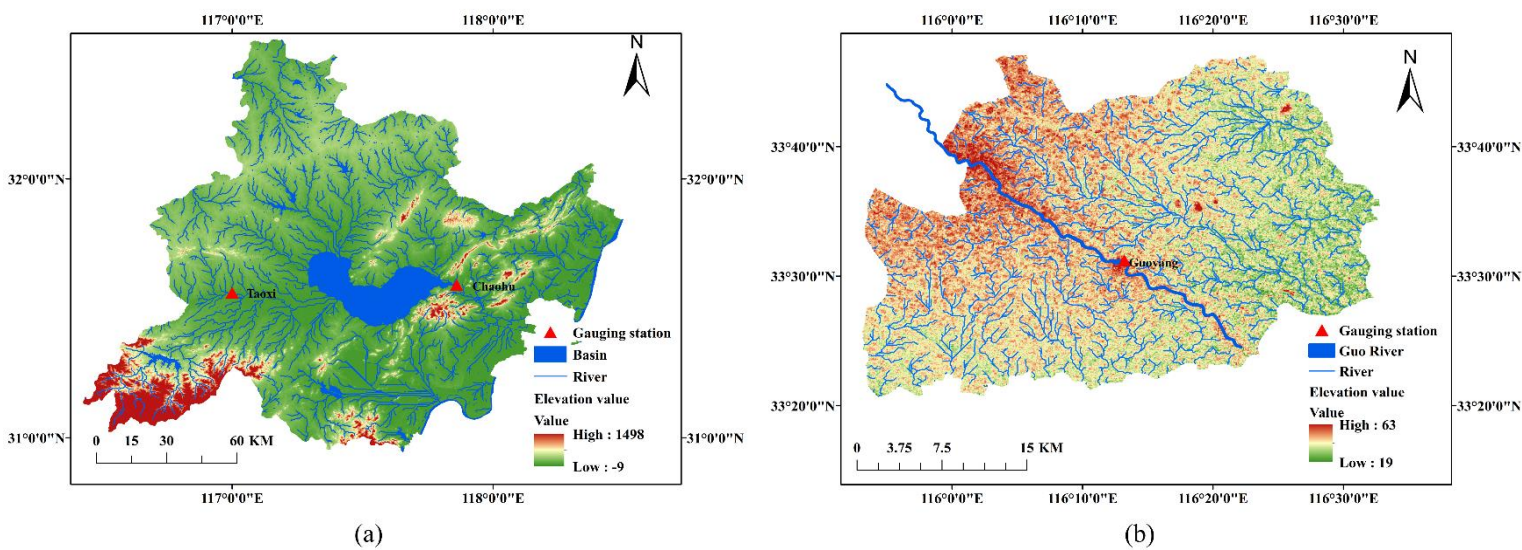

**Fig. 5.** Geographical overview of (a) Guoyang basin and (b) Chaohu basin.

# 3. Case studies

In this study, water level data from three stations (locating at Guoyang and Chaohu basins in Anhui province, China) are used to verify and validate (V&V) FSDB sampling technique. This section briefly introduces geography of Guoyang and Chaohu basins and simple characteristics of water level data from these gauging stations. Then, the statistical metrices for evaluating model performance are presented.

## 3.1 Study area

Guoyang basin is located in Guoyang County, Bozhou City, and its geographic location is shown in Fig. 5(a). The topography of the Guoyang County is gentle and the average elevation does not exceed 35 meters. The Guo river is the first major river in the Guoyang basin and the second longest tributary of the Huai River. The Guo River is located on the north bank of the Huai River



and flows into the Huai River near Huaiyuan County. Guoyang gate is located on the main stream of the East Guo River and was built in 1958.

The Chaohu basin is located in Chaohu City, and its geographic location is shown in Fig. 5(b). Chaohu basin is bounded by the Jianghuai watershed, with a long east-west and narrow north-south distribution, accounting for about 9.3% of the total area of Anhui Province. It mainly includes 16 counties and districts in Hefei, Chaohu, Luan and Anqing. The Chaohu gate, built in 1962, which is located at the entrance of Chaohu Lake into Yuxi River in the southern suburbs of Chaohu City, has played a great role in flood control, irrigation, water supply, and shipping in the Chaohu basin. Fengle River is located at the entrance of Chaohu basin and is one of the main tributaries of Chaohu Lake. Taoxi hydrological station, built in 1951, which is located at the river side of Fengle River in Taoxi Town, mainly monitoring the water level, flow, precipitation and soil moisture and other elements of Fengle River.

## 3.2 Data

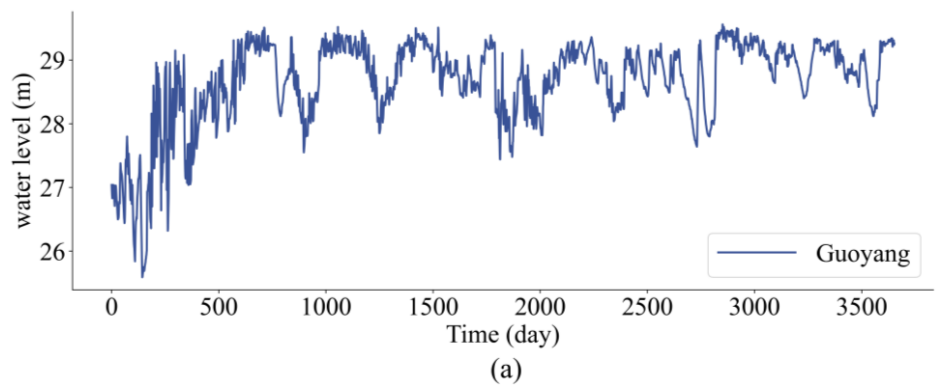

(a)

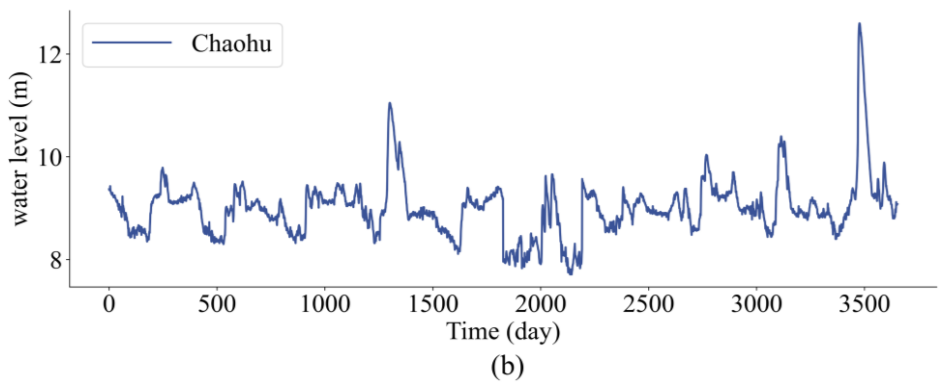

(b)



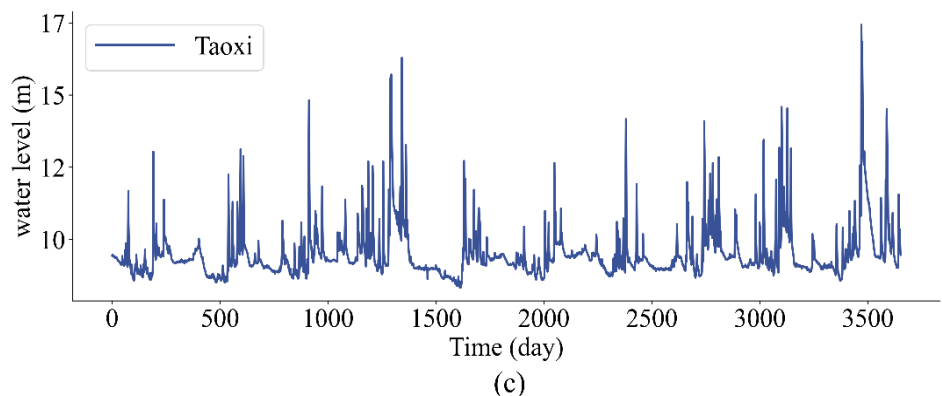

(c)

**Fig. 6.** Water level records from 2007 to 2016 at three gauging stations: (a) Guoyang station; (b) Chaohu station; (c) Taoxi station.

The available data in our study case are the upstream water level of Guoyang sluice recorded at Guoyang station, the upstream water level of Chaohu sluice recorded at Chaohu station, and the water level on the Fengle River recorded at the Taoxi station, respectively. The daily water level data cover records from 2007/1/1 to 2016/12/31, totalling 3653 days. For each gauging station, the data are first divided chronologically into calibration and validation periods. To prevent the models from overfitting, the calibration period is further divided into calibration and test subperiods, as shown in Table 2.

**Table 2**
The division of water level data of three gauging stations.

| Gauging station | Calibration period | | Validation period |
|---|---|---|---|
| | Calibration subperiod | Test subperiod | |
| Guoyang | 2007/1/1-2013/1/2 (60%) | 2013/1/3-2015/1/1 (20%) | 2015/1/2-2016/12/31 (20%) |
| Chaohu | 2007/1/1-2013/1/2 (60%) | 2013/1/3-2015/1/1 (20%) | 2015/1/2-2016/12/31 (20%) |
| Taoxi | 2007/1/1-2013/1/2 (60%) | 2013/1/3-2015/1/1 (20%) | 2015/1/2-2016/12/31 (20%) |

The water level data of the three gauging stations is shown in Fig. 6. The highest and lowest water levels at Guoyang station are 29.56m and 25.59m, respectively. The average water level is 28.66m, with small water level changes and little fluctuation. The highest, lowest, average water level and maximum water level difference at Chaohu station are 12.60m, 7.71m, 8.99m and 4.89m,



respectively. The water level at Chaohu station has obvious periodicity. The average water level at Taoxi station is only 9.50m, but the maximum water level difference was 9.1m, with strong fluctuation and periodicity. The daily water levels at all three stations show certain fluctuation and periodicity, so all three stations are characterized by non-stationarity.

## 3.3 Performance evaluation metrics

In this study, three widely used statistical evaluation criteria are used to evaluate the forecasting performance of different models, including Nash-Sutcliffe efficiency coefficient (NSE), root mean square error (RMSE) and mean absolute error (MAE).

These statistical evaluation criteria are defined as follows:

$$NSE = 1 - \frac{\sum_{t=1}^{T}(Q_t - \hat{Q}_t)^2}{\sum_{t=1}^{T}(Q_t - \bar{Q})^2} \tag{7}$$

$$RMSE = \sqrt{\frac{1}{T}\sum_{t=1}^{T}(Q_t - \hat{Q}_t)^2} \tag{8}$$

$$MAE = \frac{1}{T}\sum_{t=1}^{T}|Q_t - \hat{Q}_t| \tag{9}$$

Where T is the total number of samples, $Q_t$ and $\hat{Q}_t$ are the $t$-th observed and forecasted data, $\bar{Q}$ is the average of observed data.

The NSE is an efficiency indicator for assessing the difference between the forecasted and observed data, with a value range of $(-\infty, 1]$. The closer the value of NSE to 1, the better the model performs. The NSE is close to 0, which means that the forecasting results are close to the mean of the observed data, i.e., the overall results are reliable, but the forecasting error is large. The NSE is much less than 0, then the model is not credible. The RMSE reflects the error between the forecasted and observed data and directly reflects the accuracy of the simulation



results. The MAE is the average of the absolute error, which shows the degree of agreement between the forecasted and observed data.

# 4. Results analysis

In this section, SDB and FSDB sampling techniques are used for water level forecasting in the Guoyang and Chaohu basins. In Section 4.1, the hyperparameters of naïve XGBoost models are ~~presented~~ predefined. In Section 4.2, experimental results of above models using two different sampling techniques are compared, and then the experimental results are analyzed and summarized.

## 4.1 Hyperparameters of XGBoost models

**Table 3**
Hyperparameters of XGBoost models.

| Model | Gauging station | Hyperparameters | values | Hyperparameters | values |
|-------|-----------------|-----------------|--------|-----------------|--------|
| XGBoost | Guoyang | n_estimators | 600 | learning_rate | 0.01 |
|  |  | max_depth | 3 | min_child_weight | 1 |
|  | Chaohu | n_estimators | 450 | learning_rate | 0.01 |
|  |  | max_depth | 3 | min_child_weight | 1 |
|  | Taoxi | n_estimators | 450 | learning_rate | 0.01 |
|  |  | max_depth | 3 | min_child_weight | 1 |

The purpose of this study is to verify the applicability of the existing and proposed sampling techniques rather than using advanced decomposition methods and complex models to improve forecasting performance. By using the same baseline models, differences in forecasting performance due to inconsistent hyperparameters can be avoided, and model performance is only related to the decomposition methods and sampling techniques. In addition, the baseline models in three gauging stations are different that is to explore whether the proposed sampling technique are not limited to a specific study area and model parameters, but have broad applicability. Based on experience, the hyperparameters of baseline model at three stations are shown in Table 3.



## 4.2 Experimental results and summary

This subsection develops several SSA-models and VMD-models using the SDB and FSDB sampling techniques to determine the optimal parameters for the use of SSA and VMD. The models that have the maximum values of the NSE in the validation period are considered as the optimal models, and then these optimal models are compared with the baseline model (naïve XGBoost) to investigate the role of decomposition techniques in improving model performance. Finally, the experimental results in Section 4.2.1 and 4.2.2 are analyzed and summarized.

Two decomposition methods, SSA and VMD, are selected to pre-process the water level series. For the SSA, decomposition levels changing from 3 to 11 are tested. For the VMD, a Grid-Search (GS) is performed to determine the optimal values of the decomposition level $K$ and penalty parameter $\alpha$. By referring to Fang et al. (2019), $K$ and $\alpha$ are set to vary from (3, 100) to (11, 3000) with step sizes of 1 and 100, respectively.

### 4.2.1 Experimental results of different sampling techniques

**Table 4**
Performance of the SSA-XGBoost models developed using the SDB sampling technique

| Gauging Station | Period | DL = 3 | DL = 4 | DL = 5 | DL = 6 | DL = 7 | DL = 8 | DL = 9 | DL = 10 | DL = 11 |
|---|---|---|---|---|---|---|---|---|---|---|
| Guoyang | Calibration | 0.966 | 0.965 | 0.963 | 0.962 | 0.961 | 0.961 | 0.961 | 0.961 | 0.962 |
| | Test | 0.927 | 0.923 | 0.917 | 0.912 | 0.909 | 0.905 | 0.897 | 0.890 | 0.888 |
| | Validation | **0.894** | 0.884 | 0.882 | 0.879 | 0.880 | 0.877 | 0.875 | 0.874 | 0.875 |
| Chaohu | Calibration | 0.957 | 0.958 | 0.958 | 0.957 | 0.956 | 0.956 | 0.956 | 0.956 | 0.956 |
| | Test | 0.887 | 0.888 | 0.889 | 0.896 | 0.894 | 0.894 | 0.891 | 0.894 | 0.891 |
| | Validation | **0.836** | 0.836 | 0.832 | 0.830 | 0.828 | 0.824 | 0.820 | 0.815 | 0.813 |
| Taoxi | Calibration | 0.880 | 0.873 | 0.864 | 0.857 | 0.843 | 0.837 | 0.831 | 0.829 | 0.826 |
| | Test | 0.787 | 0.798 | 0.802 | 0.784 | 0.781 | 0.754 | 0.767 | 0.770 | 0.762 |
| | Validation | **0.797** | 0.783 | 0.796 | 0.796 | 0.792 | 0.785 | 0.783 | 0.782 | 0.781 |



**Table 5**

Performance of the SSA-XGBoost models developed using the FSDB sampling technique

| Gauging Station | Period | DL = 3 | DL = 4 | DL = 5 | DL = 6 | DL = 7 | DL = 8 | DL = 9 | DL = 10 | DL = 11 |
|---|---|---|---|---|---|---|---|---|---|---|
| Guoyang | Calibration | 0.964 | 0.964 | 0.963 | 0.963 | 0.963 | 0.963 | 0.964 | 0.962 | 0.962 |
| | Test | 0.944 | 0.944 | 0.938 | 0.938 | 0.939 | 0.940 | 0.938 | 0.936 | 0.935 |
| | Validation | 0.916 | 0.915 | 0.918 | 0.918 | 0.921 | 0.920 | 0.920 | **0.923** | 0.922 |
| Chaohu | Calibration | 0.956 | 0.960 | 0.963 | 0.965 | 0.967 | 0.969 | 0.971 | 0.973 | 0.975 |
| | Test | 0.895 | 0.903 | 0.912 | 0.921 | 0.926 | 0.931 | 0.936 | 0.940 | 0.944 |
| | Validation | 0.843 | 0.846 | 0.846 | 0.847 | 0.851 | 0.855 | 0.857 | 0.860 | **0.862** |
| Taoxi | Calibration | 0.871 | 0.861 | 0.853 | 0.847 | 0.837 | 0.834 | 0.826 | 0.821 | 0.823 |
| | Test | 0.756 | 0.752 | 0.733 | 0.735 | 0.710 | 0.704 | 0.688 | 0.682 | 0.678 |
| | Validation | 0.796 | **0.806** | 0.804 | 0.784 | 0.779 | 0.768 | 0.782 | 0.784 | 0.781 |

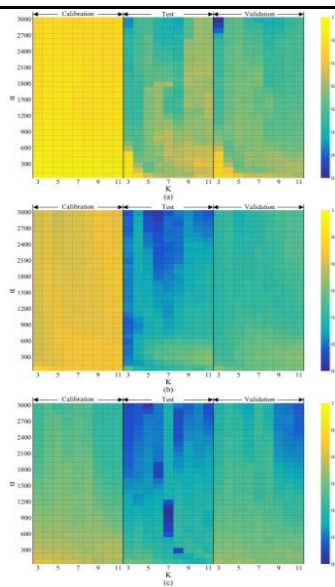

**Fig. 7.** NSE of VMD-based models developed using SDB sampling technique on calibration period, test period and validation period at (a) Guoyang station; (b) Chaohu station; and (c) Taoxi station.

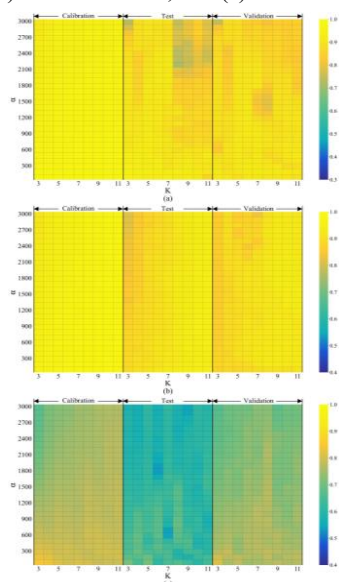

**Fig. 8.** NSE of VMD-based models developed using FSDB sampling technique on calibration period, test period and validation period at (a) Guoyang station; (b) Chaohu station; and (c) Taoxi station.



To explore the potential of SDB and FSDB sampling techniques, water level time series from three gauging stations located in two basins are used. The model performance is shown in Tables 4-5 and Figs. 7-8.

Tables 4-5 show the models performance using SSA decomposition method. As shown in Table 4, when the decomposition level (DL) is 3, the models developed by SDB sampling technique (SDB-models) achieves the best performance at all three stations, and the values of the NSE are 0.894, 0.836 and 0.797, respectively. However, the forecasting performance tends to decrease as the DL increases. This may be because the model has prediction error for each component, and as the number of components increases, the prediction errors get accumulated, which eventually leads to a decline in the forecasting performance. Table 5 shows the performance of the models developed by the FSDB sampling technique (FSDB-models) at different DLs at three stations. Unlike SDB-models, the FSDB-models do not achieve the maximum NSE values at the lowest DL. This may be because the model can obtain accurate prediction results for each component at lower DLs. As the DL increases, the model's prediction performance tends to increase as the pattern of each component becomes simpler and more independent. However, as the DL continues to increase, the performance of the models gradually declines. On the one hand, this may be because some of the components have little contribution to the original time series when the DL is large, and can be regarded as noise. The models have weak ability to predict the noise with randomness, which leads to the decline of the overall forecasting performance. On the other hand, time series are often a superposition or coupling of trend, periodicity, fluctuation, etc. The over-decomposition may lead to unclear patterns of the components and the models are unable to extract valid information from them. When the decomposition levels are 10, 11 and 4, the maximum NSE values of the FSDB-models at three



stations are 0.923, 0.862 and 0.806, respectively, which are 3.2%, 3.1% and 1.1% higher than SDB-models. In addition, the FSDB-models show better forecasting performance in most cases except for few DLs at Taoxi station.

Figs. 7-8 show the models performance using VMD decomposition method. As shown in Fig. 7, the SDB-models exhibit better performance when the decomposition level $K$ or the penalty parameter $\alpha$ is small. The maximum NSE values for the three studied stations are 0.859, 0.726 and 0.755 when the combinations of $K$ and alpha are (3, 100), (11,300) and (3, 100), respectively. The performance of SDB-models tends to decrease as the values of $K$ and alpha increase. On the one hand, this may be due to the fact that a too-large value of $K$ will cause the original series to be over-decomposed, where several IMFs share one center frequency, resulting in mode mixing or frequency mixing. On the other hand, A too-large value of alpha may result in smaller bandwidths of the IMFs and thus loss of some information of the original series. It is observed that there is an obvious overfitting of SDB-models. Fig. 8 shows the forecasting performance of the FSDB-models using the VMD decomposition method. When the combinations of $K$ and $\alpha$ are (3, 100), (11, 1600), and (3, 100), the maximum values of the NSE are obtained as 0.914,0.935 and 0.808, respectively, which are 6.4%, 28.8%, and 7.0% higher than those of the SDB-models. The FSDB-models exhibit high forecasting accuracy for various combinations of $K$ and alpha, and no significant overfitting is observed.



## 4.2.2 Experiment results of decomposition-based models and benchmark models

In Section 4.2.1, several SSA-models and VMD-models have been developed using the SDB and FSDB sampling techniques due to a lack of a reliable support to determine the optimal parameters for the use of SSA and VMD. The models with the maximum values of the NSE in the validation period are selected as the optimal models. A comparison is conducted between these optimal models and the benchmark (the naïve XGBoost) models in this section.

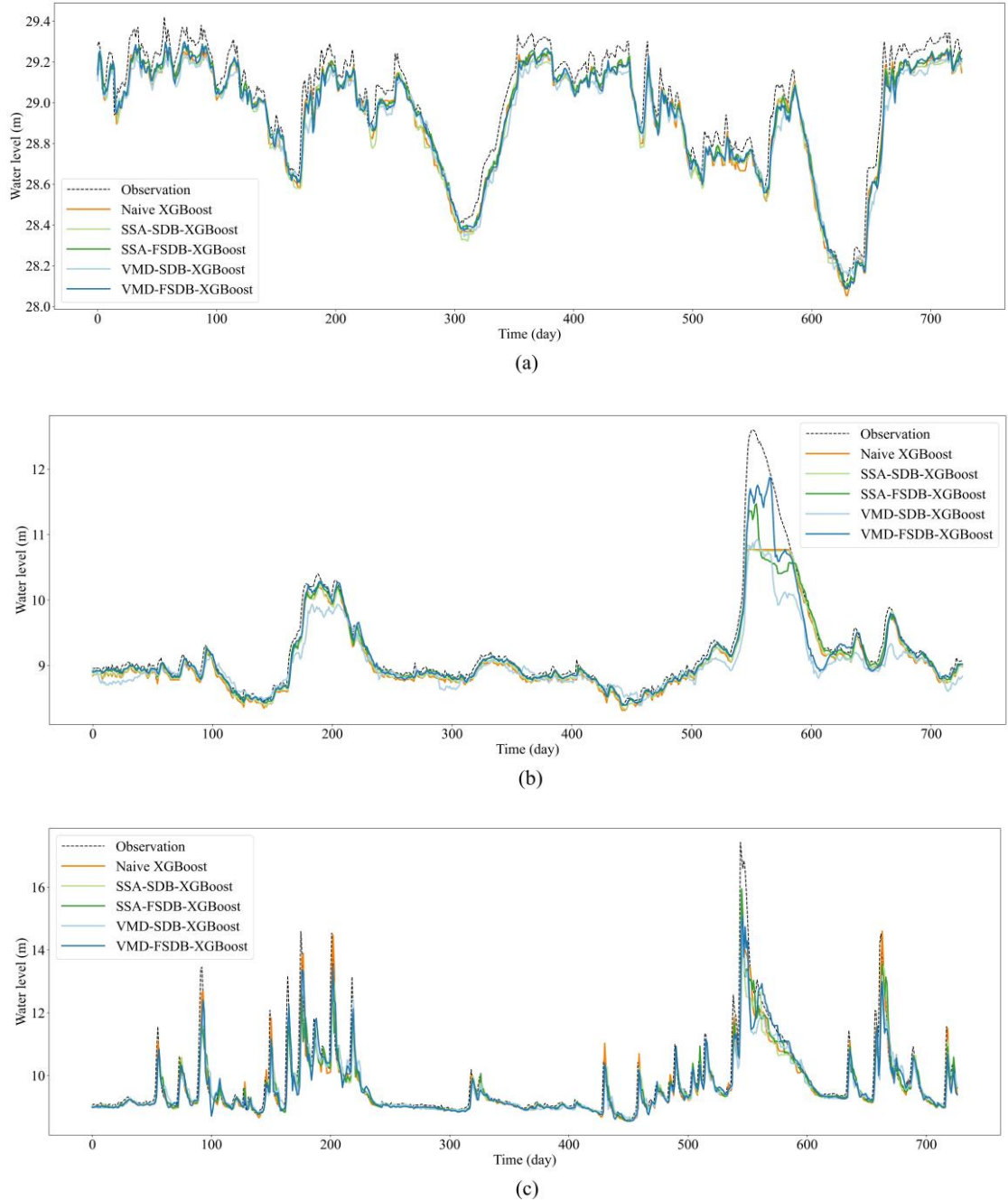

(a)

(b)

(c)

**Fig. 9.** Comparison between observed and forecast water level in the validation period. (a) Guoyang station; (b) Chaohu station; (c) Taoxi station.

Fig. 9 clearly shows that all these models provide good performance in tracking the fluctuation in water level series of three stations, which proves the overall feasibility of the forecasting models. As shown in Table 6 and Fig. 10, at two of the three stations, the decomposition-based models exhibit better performance than the naïve XGBoost models and the optimal models are developed using FSDB sampling technique. At Guoyang station, the SSA-FSDB-XGBoost model has the maximum NSE which is 0.923 and minimum RMSE, MAE which are 0.081 and 0.064 that are 1.54% higher and 7.95%, 15.79% lower than those of the naïve XGBoost model, respectively. At Chaohu station, the performance improvement of hybrid models is more obvious. The NSE, RMSE and MAE of VMD-FSDB-XGBoost are 0.935, 0.195 and 0.108 which are 11.6% higher and 36.7%, 29.4% lower than those of the naïve XGBoose model, respectively.

However, it should be noted that the naïve XGBoost model has a higher forecasting accuracy than decomposition-based hybrid models at Taoxi station. This could be because the daily water level series at Taoxi changed frequently and with a large variation and strong non-stationary from 2007 to 2016, whereas those of the Guoyang and Chaohu stations did not present significant non-stationary behavior.



**Table 6**

Performance comparison between the baseline model and the optimal models developed using the SDB and FSDB sampling techniques.

| Gauging station | Model | Calibration | | | Test | | | Validation | | |
|---|---|---|---|---|---|---|---|---|---|---|
| | | NSE | RMSE | MAE | NSE | RMSE | MAE | NSE | RMSE | MAE |
| Guoyang | Naïve XGBoost | 0.965 | 0.141 | 0.105 | 0.943 | 0.109 | 0.083 | 0.909 | 0.088 | 0.076 |
| | SSA-SDB-XGBoost | 0.966 | 0.136 | 0.106 | 0.927 | 0.122 | 0.091 | 0.894 | 0.094 | 0.081 |
| | SSA-FSDB-XGBoost | 0.962 | 0.146 | 0.108 | 0.936 | 0.116 | 0.080 | **0.923** | 0.081 | 0.064 |
| | VMD-SDB-XGBoost | 0.947 | 0.174 | 0.133 | 0.898 | 0.146 | 0.108 | 0.859 | 0.110 | 0.090 |
| | VMD-FSDB-XGBoost | 0.964 | 0.143 | 0.107 | 0.943 | 0.120 | 0.080 | 0.914 | 0.086 | 0.070 |
| Chaohu | Naïve XGBoost | 0.950 | 0.115 | 0.096 | 0.873 | 0.109 | 0.098 | 0.838 | 0.308 | 0.153 |
| | SSA-SDB-XGBoost | 0.957 | 0.106 | 0.086 | 0.887 | 0.102 | 0.087 | 0.836 | 0.308 | 0.144 |
| | SSA-FSDB-XGBoost | 0.975 | 0.081 | 0.054 | 0.944 | 0.072 | 0.051 | 0.862 | 0.284 | 0.115 |
| | VMD-SDB-XGBoost | 0.837 | 0.207 | 0.160 | 0.726 | 0.161 | 0.125 | 0.726 | 0.400 | 0.229 |
| | VMD-FSDB-XGBoost | 0.970 | 0.088 | 0.057 | 0.945 | 0.072 | 0.051 | **0.935** | 0.195 | 0.108 |
| Taoxi | Naïve XGBoost | 0.887 | 0.272 | 0.135 | 0.816 | 0.306 | 0.152 | **0.834** | 0.506 | 0.226 |
| | SSA-SDB-XGBoost | 0.880 | 0.259 | 0.131 | 0.787 | 0.298 | 0.155 | 0.797 | 0.530 | 0.251 |
| | SSA-FSDB-XGBoost | 0.861 | 0.301 | 0.138 | 0.752 | 0.355 | 0.161 | 0.806 | 0.547 | 0.244 |
| | VMD-SDB-XGBoost | 0.813 | 0.349 | 0.164 | 0.696 | 0.393 | 0.185 | 0.755 | 0.614 | 0.293 |
| | VMD-FSDB-XGBoost | 0.844 | 0.319 | 0.150 | 0.709 | 0.384 | 0.177 | 0.808 | 0.544 | 0.251 |

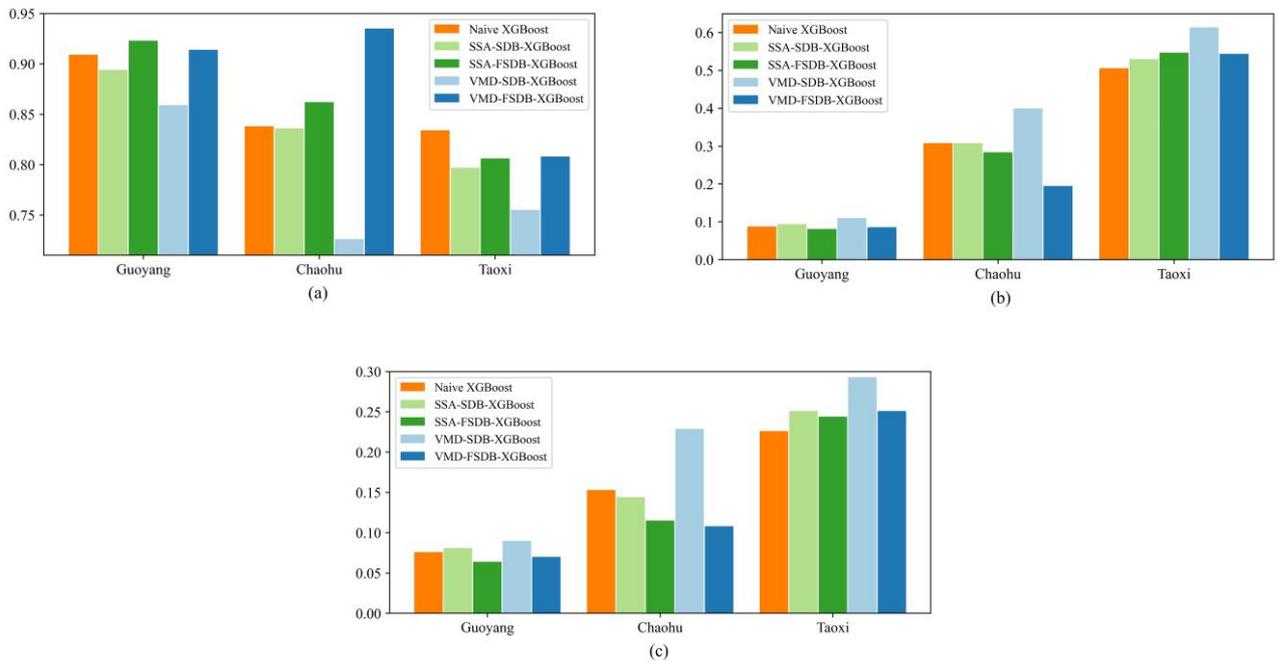

**Fig. 10.** (a) NSE, (b) RMSE and (c) MAE of XGBoost based on different sampling techniques and decomposition methods in validation period at three stations.



### 4.2.3 Discussion of experiment results

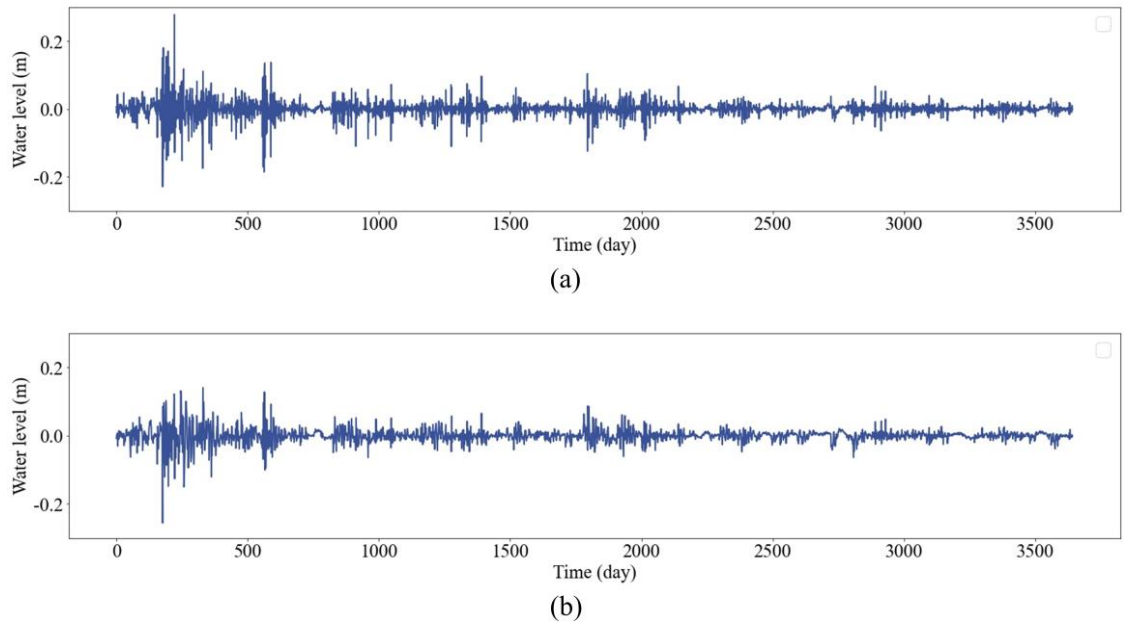

**Fig. 11.** Differences between original series and VMD reconstructed series using (a) SDB sampling technique and

(b) FSDB sampling technique when $K$ and alpha are (3, 100) at Guoyang station.

In subsection 4.2.1 and 4.2.2, the model performances in the combination of different sampling

techniques, different decomposition methods and different gauging stations are compared. This

subsection further analyzes the reasons for these differences in model performance.

In most cases, results of hybrid models using FSDB sampling technique are better than those

using SDB sampling technique. The reason for this is the different way of obtaining response

variables. When SDB sampling technique decomposes the entire time series, the decomposition of

the current value is affected by future data, resulting in biased response variables. This causes the

models' forecasting results to be biased towards the wrong 'true values' and thus deviate significantly

from the actual true values. However, the FSDB sampling technique obtains response variables of

the previous sample while extracting explanatory variables of the current sample. This cleverly

avoids bias due to future data.



Difference also appears in the model performance using different decomposition method, showing in Figs. 7-8 and Table 6 (detailed introduction in section 4.2.1). Compared with SSA-based models, the performance differences between FSDB and SDB sampling techniques is more apparent for VMD-based models. The reason for this is the high proportion of reconstruction error. Usually, there are two different kinds of errors in VMD-based forecasting model, i.e., forecasting error and reconstruction error. Forecasting errors mainly refer to the baseline model performance, which depends on the forecasting ability of ML models. Reconstruction errors represent the difference between the reconstructed data, i.e., the sum of all components obtained by VMD, and original data. Due to the reconstruction error, the sum of response variables obtained from the decomposition is not the same as original data. This means even if the model accurately forecasts each response variable, the final forecast is different with original data. Therefore, reducing the reconstruction error means that the reconstructed series is as much as possible the same as the original series. As shown in Fig. 11, the SDB sampling technique extracts the affected response variables by decompose the overall series, which lead to biased reconstructed series and exacerbates the reconstruction error. In contrast, the proposed FSDB sampling technique reduces the reconstruction error by extracting response variables of all samples in a fully stepwise decomposition way. Therefore, the FSDB sampling technique substantially improves the forecasting performance of the VMD-based models. As for SSA-based models, there are only forecasting errors because the sum of all components obtained by SSA is equal to original series.

In addition, the experimental results in Section 4.2.2 shows that the models exhibited different performance in different study areas. The decomposition-based models exhibit better forecasting performance at Guoyang and Chaohu stations, while the naive XGBoost model has higher



forecasting accuracy at Taoxi station. The reason for the exceptional results could be the obvious non-stationary water levels at Taoxi station. The daily water level series at Guoyang and Chaohu stations has regular periodic variation, small fluctuations and weak non-stationary. In contrast, the water level at Taoxi station changed frequently and with a large variation, showing an obvious non-stationary behavior. Therefore, it can be inferred that even with the introduction of the series decomposition method, the non-stationary of water level may be a major obstacle to achieving satisfactory model performance, as confirmed by Fang et al. (2019). In most cases, decomposition methods such as the SSA and VMD can play positive roles in improving model performance. However, in real-world applications, it is necessary to conduct specific studies for specific problems to determine whether the decomposition methods are applicable. When decomposition methods can improve the model performance, the FSDB is a effective sampling technique.

# 5. Conclusions and future works

In this paper, a Fully Stepwise Decomposition-Based sampling technique considering response variables correction, i.e., FSDB, is proposed to improve the accuracy and practicality of the decomposition-based hybrid model. The FSDB sampling technique can generate these response variables unaffected by future data to avoid biased forecasting targets. To verify and validate this new technique, experiments of daily water level forecasting for two types of hybrid models using SDB and FSDB sampling techniques are implemented at three gauging stations in Guoyang and Chaohu basins in China. Results show that, SSA and VMD decomposition-based model using FSDB sampling technique, compared with those using SDB sampling techniques, have a good promoting effect on the prediction result. Because FSDB sampling technique strictly excludes information on



future water level. Meanwhile, SDB sampling technique decomposes the overall series to extract response variables, which introduces additional information on future water levels and leads to biased forecasting targets. The forecasting performance of these decomposition-based models are also compared with those of the naïve XGBoost models. Results show that the decomposition-based models using the FSDB sampling technique exhibit the best forecasting performance at Guoyang and Chaohu stations, but the performance is not as good as the naïve XGBoost models at Taoxi station. The reason for this exception is that, the non-stationarity of water level data at Taoxi station is so acute that can be not erased by the decomposition method. It is also meaningful to develop a better decomposition method in the future. In summary, the new FSDB sampling technique is of practical significance, and is proved to be more accurate than current advanced sampling technique, which can be extended to other application of decomposition-based hybrid model.

In the future, we would like to consider about the problem of slow calculation speed of existing decomposition-based model, which makes real-time application.

# Acknowledgements

This work is supported by Hefei Key Generic Technology R&D and Major Achievements Engineering Project (2021GJ012). The authors would like to thank hydrographic office of Fuyang city and Chaohu Research Institute in China especially Juan Tian for her helpful discussions and supply of rich hydrological data. The authors also would like to thank members of Anhui Jinhaidier Information Technology Co. Ltd., Hefei, China, for their helpful discussions and supports.